\def\paperTitle{\modelbold: Self-Supervised Representation Learning from Images and Video with Contrastive Masked Autoencoders}

\def\authorBlock{
    Jefferson Hernandez$^{1}$, Ruben Villegas$^{2}$, Vicente Ordonez$^{1}$ 
}
\def\instBlock{
    \vspace{-0.1in}
    $^{1}$Rice University, $^{2}$Google DeepMind\\
    {\tt\small \{jefehern, vicenteor\}@rice.edu, rubville@google.com} \\
    Code \& Weights: \url{https://github.com/jeffhernandez1995/ViC-MAE}
}

\documentclass[runningheads]{llncs}

\usepackage[mobile]{eccv}

\usepackage{eccvabbrv}

\usepackage{graphicx}	
\usepackage{amsmath}	
\usepackage{amssymb}
\usepackage{wrapfig}
\usepackage{booktabs}
\usepackage{makecell} 
\usepackage{microtype}
\usepackage{epsfig}
\usepackage{caption}
\usepackage{float}
\usepackage{placeins}
\usepackage{color, colortbl}
\usepackage{stfloats}
\usepackage{enumitem}
\usepackage{tabularx}
\usepackage{xstring}
\usepackage{multirow}
\usepackage{xspace}
\usepackage{url}
\usepackage{subcaption}
\usepackage{xcolor}
\usepackage[hang,flushmargin]{footmisc}

\usepackage{bbm}
\usepackage{enumitem}
\usepackage{pifont}
\usepackage{makecell} % Load the makecell package
\usepackage{array}
\usepackage[utf8]{inputenc} % allow utf-8 input
\usepackage[T1]{fontenc}    % use 8-bit T1 fonts
\usepackage{amsfonts}       % blackboard math symbols
\usepackage{nicefrac}       % compact symbols for 1/2, etc.
\usepackage{microtype}      % microtypography
\usepackage{calc}
\usepackage{setspace}

\newcommand\vicmaefontbold[1]{\smash{{\usefont{T1}{minabold}{m}{n}#1}}}

\newcommand\vicmaefont[1]{\smash{{\usefont{T1}{mina}{m}{n}#1}}}

\usepackage{listings}
\usepackage[ruled]{algorithm2e}

\usepackage{tikz}
\def\checkmark{\tikz\fill[scale=0.4](0,.35) -- (.25,0) -- (1,.7) -- (.25,.15) -- cycle;}
\usepackage[accsupp]{axessibility}  % Improves PDF readability for those with disabilities.

\newcommand{\supp}{supplemental material\xspace}

\newcommand{\R}[1]{{%
    \textbf{%
        \ifstrequal{#1}{1}{\textcolor{red}{R#1}}{%
        \ifstrequal{#1}{2}{\textcolor{blue}{R#1}}{%
        \ifstrequal{#1}{3}{\textcolor{magenta}{R#1}}{%
        \ifstrequal{#1}{4}{\textcolor{teal}{R#1}}{%
                           \textcolor{cyan}{R#1}%
        }}}}%
    }%
}}

\newcommand{\model}{\vicmaefont{ViC-MAE}\xspace}

\newcommand{\modelbold}{\vicmaefontbold{ViC-MAE}\xspace}

\makeatletter
\usepackage{hyperref}

\usepackage[capitalize]{cleveref}
\crefname{section}{Sec.}{Secs.}
\crefname{table}{Table}{Tables}
\crefname{figure}{Fig.}{Figs.}

\usepackage{orcidlink}

\begin{document}

\title{\paperTitle}

\titlerunning{Visual Contrastive Masked Autoencoders}

\author{\authorBlock}
\authorrunning{Hernandez, Villegas and Ordonez}
\institute{\instBlock}

\maketitle

\begin{abstract} 
    We propose \model, a model that combines both Masked AutoEncoders (MAE) and contrastive learning. \model is trained using a global representation obtained by pooling the local features learned under an MAE reconstruction loss and using this representation under a contrastive objective across images and video frames.
    We show that visual representations learned under \model generalize well to video and image classification tasks. 
    Particularly, \model obtains state-of-the-art transfer learning performance from video to images on Imagenet-1k compared to the recently proposed OmniMAE by achieving a top-1 accuracy of 86\% (+1.3\% absolute improvement) when trained on the same data and 87.1\% (+2.4\% absolute improvement) when training on extra data. At the same time, \model outperforms most other methods on video benchmarks by obtaining 75.9\% top-1 accuracy on the challenging Something something-v2 video benchmark.
    When training on videos and images from diverse datasets, our method maintains a balanced transfer-learning performance between video and image classification benchmarks, coming only as a close second to the best-supervised method. 
    \vspace{-0.2in}

\end{abstract}

\section{Introduction}
\label{sec:intro}

Recent advances in self-supervised visual representation learning have markedly improved performance on image and video benchmarks~\cite{chen2020simple, he2020momentum, caron2021emerging, he2022masked}. This success has been mainly driven by two approaches: Joint-embedding methods, which encourage invariance to specific transformations—either contrastive~\cite{chen2020simple, he2020momentum, caron2021emerging} or negative-free~\cite{chen2021exploring, bardes2021vicreg}, and masked image modeling which works by randomly masking out parts of the input and forcing a model to predict the masked parts with a reconstruction loss~\cite{bao2021beit, he2022masked, feichtenhofer2022masked, wei2022masked}. These ideas have been successfully applied to both images and video.

\begin{figure}[t!]
    \centering
    \includegraphics[width=0.7\linewidth]{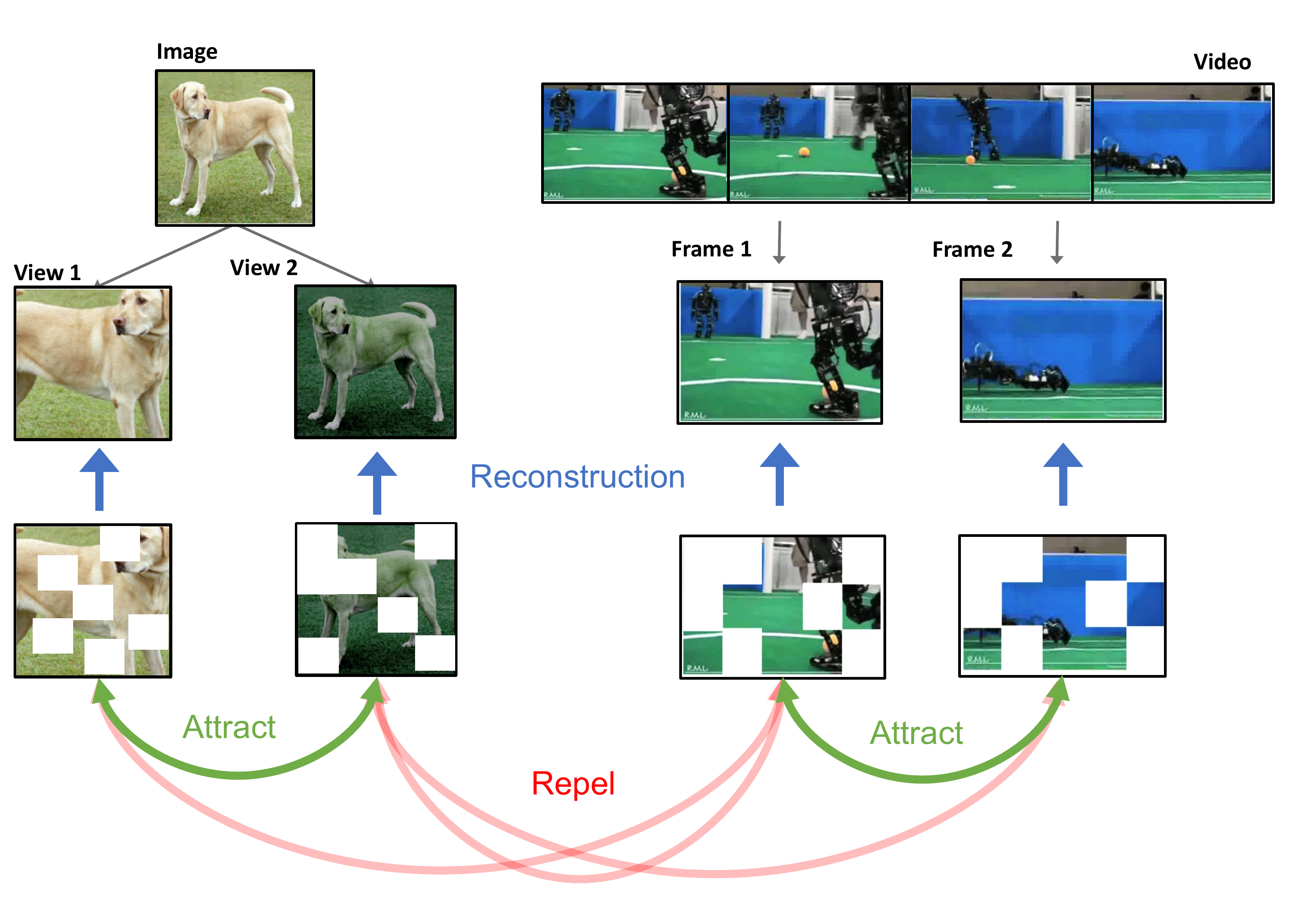}
    \vspace{-0.2in}
    \caption{\model operates over video frames and images using masked image modeling at the image and frame level and contrastive learning at the temporal level for videos and under image transformations for images. Our model represents a strong backbone for both image and video tasks.
    }
    \label{fig:vic-mae_idea}
    \vspace{-0.25in}
\end{figure}

Self-supervised techniques for video representation learning have resulted in considerable success, yielding powerful features that perform well across various downstream tasks~\cite{feichtenhofer2022masked, wei2022masked, qian2021spatiotemporal, feichtenhofer2021large}. Leveraging image-based models to enhance video feature representations has gained widespread adoption, evidenced by significant advancements in robust video representations~\cite{liu2022video, arnab2021vivit, li2022mvitv2}.  The reverse---{\em video-to-image} transfer learning---has not been as successful. This imbalance underscores a nuanced challenge within multimodal learning, and it is not clear how to integrate different modalities. Furthermore, attempts to combine these modalities often result in diminished performance, necessitating tailored adjustments to the underlying architectures or converting one modality (images) into another (repeating images to simulate a video).
Learning from video should also yield good image representations since videos naturally contain complex changes in pose, viewpoint, and deformations, among others. These variations can not be simulated through the standard image augmentations used in joint-embedding methods or masked image modeling methods. In this work, we propose a {\bf Vi}sual {\bf C}ontrastive {\bf M}asked {\bf A}uto{\bf E}ncoder (\model), a model that learns from both images and video through self-supervision, instead treating short videos as the different views of the same representation, diverging from previous works~\cite{girdhar2022omnivore, girdhar2023omnimae}. On transfer experiments, our model also improves {\em video-to-image} transfer performance while maintaining performance on video representation learning.

Prior work has successfully leveraged self-supervision for video or images separately using either contrastive learning (\ie~Gordon~\etal~\cite{gordon2020watching}), or masked image modeling (\ie~Feichtenhofer~\etal~\cite{feichtenhofer2022masked}). \model seeks to leverage the strength of contrastive learning and masked image modeling and seamlessly incorporate images. While trivially this has been done by repeating the image to simulate a still video, \model achieves this in the opposite way, treating frames sampled within short intervals (\eg~$~1\text{sec}$) as an additional form of temporal data augmentation. 
Our method uses contrastive learning to align representations across both time-shifted frames and augmented views, and masked image modeling for single video frames or images to encourage learning local features. Diverging from methods that only use a $[\text{CLS}]$ token as a global feature, our model aggregates local features using a global pooling layer followed by a contrastive loss to enhance the representation further. This structure is built upon the foundation of the Vision Transformer~(ViT) architecture~\cite{dosovitskiy2020image}, which has become a standard for masked image modeling methods. 

\begin{figure*}[ht!]
    \centering
    \includegraphics[width=\linewidth]{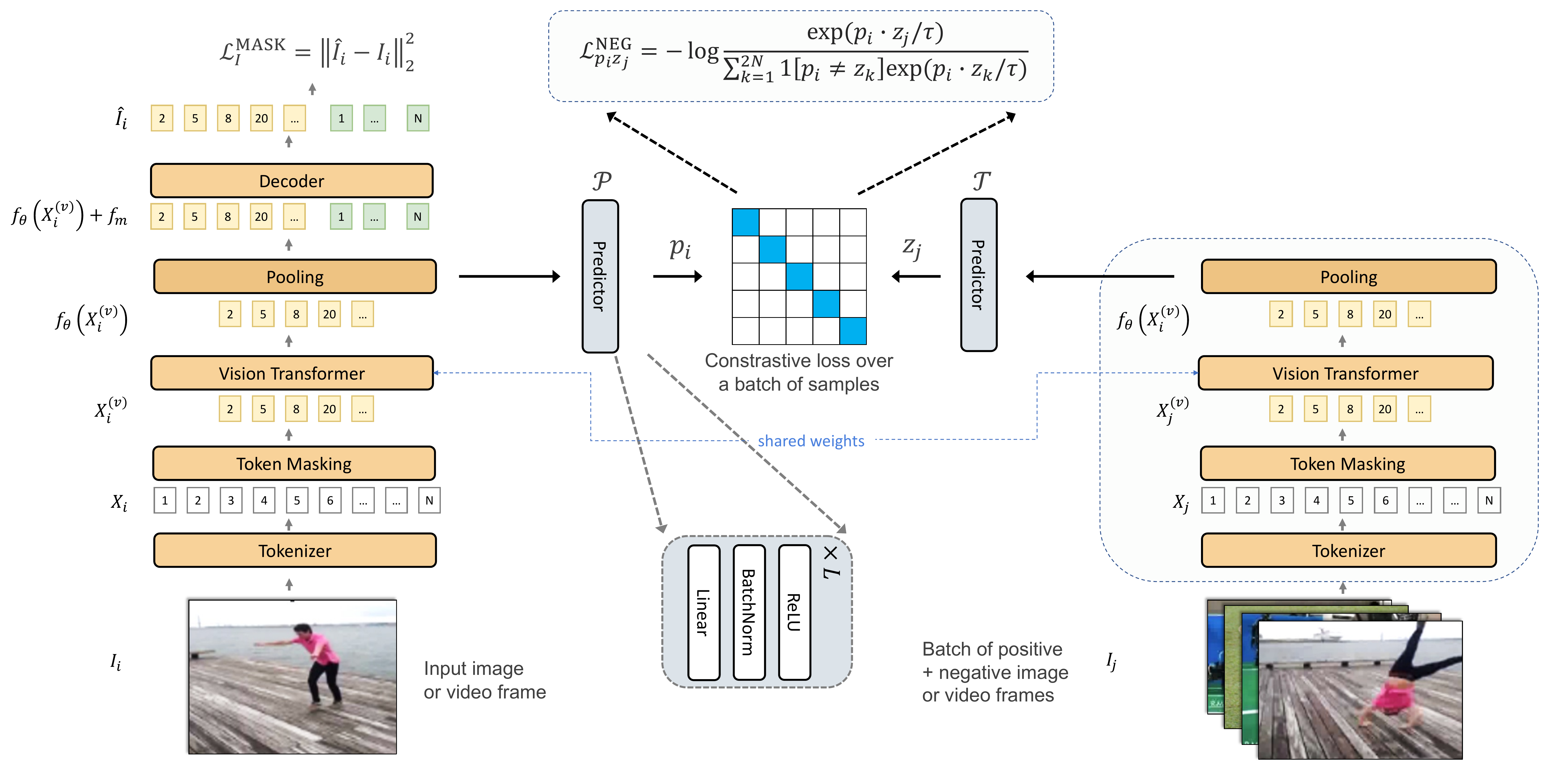}
    \vspace{-0.15in}
    \caption{\textbf{\model} inputs two distant frames from a video or two different views of an image within the same batch using a siamese backbone (shared weights), and randomly masks them, before passing them through a ViT model which learns a representation of local features using masked image modeling. A global representation of the video is then constructed by global pooling of the local features learned by the ViT model trained to reconstruct individual patches using an $\ell_2$ loss. A standard predictor and a target encoder are used with a contrastive loss. Our use of an aggregation layer before the predictor network aids in avoiding the collapse of the learned global representations.}
    \label{fig:vic-mae}

\vspace{-0.2in}
\end{figure*}

Closely related to our work is the recently proposed \mbox{OmniMAE}~\cite{girdhar2023omnimae} which also aims to be a self-supervised model that can serve as a foundation for image and video downstream tasks. While our experimental evaluations compare \model favorably especially when relying on the ViT-L architecture (86\% top-1 accuracy on Imagenet vs 84.7\%, and 86.8\% top-1 accuracy on Kinetics-400 vs 84\%), there are also some fundamental differences in the methodology. OmniMAE relies exclusively on masked image modeling and treats images as videos, while \model samples frames more sparsely, treating videos within a short time span as the same view. \model, leads to reduced training times than video-masked models, while it demands more resources than a basic MAE (which processes 49 visual tokens at a 75\% masking rate), it is more efficient (handling 98 tokens at the same rate) than heavier models like OmniMAE or ST-MAE (157 tokens at 90\% rate). Surprisingly with these simplications, \model works and achieves high performance on video and image tasks, learning effective temporal representations when finetuning on video. Ultimately, we consider our contributions to be orthogonal and could potentially be integrated to achieve further gains. 

Our main empirical findings can be summarized as follows: 
(i) Treating short videos as augmented views, and then finetuning on regular videos or images yields stronger performance than treating images as videos, while the end models still retain temporal representations,
(ii) training with large frame gaps (approx 1.06 seconds) between sampled frames enhances classification performance, providing the kind of strong augmentation that joint-embedding methods typically require, 
(iii) including negative pairs in training outperforms negative-free sample training,\footnote{See \supp for an evaluation of what we tried and did not work when combining negative-free methods with masked image modeling} aligning with other methods that have been successful in {\em video-to-image} evaluations, and  
(iv) training with strong image 
transformations as augmentations is necessary for good performance on images.

Our contributions are as follows: (1) We introduce \model, which combines contrastive learning with masked image modeling that works on videos and images by treating short videos as temporal augmentations, unlike previous works; (2) When \model is trained only on videos, we achieve state-of-the-art {\em video-to-image} transfer learning performance on the ImageNet-1K benchmark and state-of-the-art self-supervised performance for video classification on SSv2~\cite{goyal2017something}; and (3) We demonstrate that \model achieves superior transfer learning performance across a wide spectrum of downstream image and video classification tasks, outperforming baselines trained only with masked image modeling. 
Our source code and model checkpoints are available \href{https://github.com/jeffhernandez1995/ViC-MAE}{here}. 

\section{Related Work}
\label{sec:related}
\vspace{-0.1in}
Our work is related to various self-supervised learning strategies focusing on video and image data, especially in enhancing image representation through video. 

%\vspace{-0.01in}
\noindent
\textbf{Self-supervised Video Learning.}
Self-supervised learning exploits temporal information in videos to learn representations aiming to surpass those from static images by designing pretext tasks that use intrinsic video properties such as frame continuity~\cite{srivastava2015unsupervised, walker16, vondrick16, mathieu16, lotter17, diba17}, alongside with object tracking \cite{agrawal2015learning, wang15, pathak2017learning, wang19}. Contrastive learning approaches on video learn by distinguishing training instances using video temporality \cite{bardes2021vicreg, chen2021exploring, wu2021contrastive, xu2021rethinking, gordon2020watching, parthasarathy2022self}. Recently, Masked Image Modeling (MIM) has used video for pre-training either using the standard design ~\cite{he2022masked} or an asymmetrical siamese design that predicts future masked frames conditioned on present unmasked frames~\cite{NEURIPS2023_7ffb9f1b}; aiding in transfer learning for various tasks~\cite{feichtenhofer2022masked,wei2022masked,tong22}.
Our approach uniquely integrates contrastive learning and masked image modeling into a single pre-training framework suitable for image and video downstream applications.

\vspace{0.01in}
\noindent
\textbf{Learning {\em video-to-image} representations.}
Several previous models trained only on images have demonstrated remarkable {\em image-to-video} adaptation~\cite{liu2022video, arnab2021vivit, li2022mvitv2}. However, static images lack the dynamism inherent to videos, missing motion cues and camera view changes. In principle, this undermines image-based models for video applications. Recent work has leveraged video data to learn robust image representations to mitigate this. For instance, VINCE~\cite{gordon2020watching} shows that natural augmentations found in videos could outperform synthetic augmentations. %and even ImageNet-1k pre-training. 
VFS~\cite{xu2021rethinking} uses temporal relationships to improve results on static image tasks. CRW~\cite{wu2021contrastive} employs cycle consistency for inter-video image mapping, allowing for learning frame correspondences. ST-MAE~\cite{feichtenhofer2022masked} shows that video-oriented masked image modeling can benefit image-centric tasks. VITO~\cite{parthasarathy2022self} develops a technique for video dataset curation to bridge the domain gap between video and images.

\vspace{0.01in}
\noindent
\textbf{Learning general representations from video and images.}
Research has progressed in learning from video and images, adopting supervised or unsupervised approaches.  The recently proposed TubeViT~\cite{piergiovanni2022rethinking} uses sparse video tubes for creating visual tokens across images and video. \mbox{OMNIVORE}~\cite{girdhar2022omnivore} employs a universal encoder for multiple modalities with specific heads for each task. PolyViT~\cite{likhosherstov2021polyvit} additionally trains with audio data, using balanced task-training schedules. Expanding on the data modalities, ImageBind~\cite{girdhar2023imagebind} incorporates audio, text, and various sensor data, with tailored loss functions and input sequences to leverage available paired data effectively. In self-supervised learning, BEVT~\cite{wang2022bevt} adopts a BERT-like approach for video, finding benefits in joint pre-training with images. OmniMAE~\cite{girdhar2023omnimae} proposes masked autoencoding for joint training with video and images. 
OmniVec~\cite{srivastava2024omnivec} extends the datasets using in \mbox{OMNIVORE}, creates new task training policies, and adds masked autoencoding as an auxiliary task to learn from multiple modalities.
\model learns from video and image datasets without supervision by combining masked image modeling and contrastive learning.

\vspace{0.02in}
\noindent
\textbf{Combining contrastive methods with masked image modeling.}
Contrastive learning combined with masked image modeling has been recently investigated. MSN~\cite{assran2022masked} combines masking and augmentations for efficient contrastive learning, using entropy maximization instead of pixel reconstruction to avoid representational collapse, achieving notable few-shot classification performance on ImageNet-1k. CAN~\cite{mishra2022simple} uses a framework that combines contrastive and masked modeling, employing a contrastive task on the representations from unmasked patches and a reconstruction plus denoising task on visible patches. C-MAE~\cite{huang2022contrastive} uses a Siamese network design comprising an online encoder for masked inputs and a momentum encoder for full views, enhancing the discrimination power of masked autoencoders which usually lag in linear or KNN evaluations. C-MAE-V~\cite{lu2023cmae} adapts C-MAE to video, showing improvements on Kinetics-400 and Something Something-v2. MAE-CT~\cite{lehner2023contrastive} leverages a two-step approach with an initial masked modeling phase followed by contrastive tuning on the top layers, improving linear classification on masked image modeling-trained models. Our \model sets itself apart by effectively learning from images and videos within a unified training approach, avoiding the representational collapse seen in C-MAE through a novel pooling layer and utilizing dual image crops from data augmentations or different video frames to improve modality learning performance.

\section{Method}
\label{sec:method}

We propose \model for feature learning on video and images, which works using contrastive learning at the temporal level (or augmentations on images) and masked image modeling at the image level.

\subsection{Background}
We provide below some background terminology and review of closely related methods that we build upon.

\vspace{0.02in}
\noindent
\textbf{Masked image modeling.}
This approach provides a way to learn visual representations in a self supervised manner. These methods learn representations by first masking out parts of the input and then training a model to fill in the blanks using a simple reconstruction loss. To do this, these methods rely on an encoder $f_{\theta}$ that takes the non-masked input and learns a representation $x$, such that a decoder $d_{\phi}$ can reconstruct the masked part of the input. More formally, let $x$ be the representation learned by the encoder for masked image $I$ with mask $M$ such that $f_{\theta}(I \odot M)$. A decoder $d$ is then applied to obtain the first loss over masked and unmasked tokens $d_{\phi}(x)$. This defines the following reconstruction loss which is only computed over masked tokens:
\begin{equation}
    \begin{aligned}
    \mathcal{L}^{\text{MASK}}_{I} = \left \|d_{\phi}(f_{\theta}(I \odot M)) \odot (1 -M) %\\
    - I\odot (1-M ) \right \|_2.
    \end{aligned}
    \label{eq: recons}
\end{equation}

\vspace{0.02in}
\noindent
\textbf{Contrastive learning.}
In common image-level contrastive methods, learning with negatives is achieved by pushing the representation of the positive pairs (different augmented views of the same image) to be close to each other while pulling the representation of negative pairs further apart. More formally, let $I$ and $I'$ be two augmented views of the same image. Contrastive learning uses a siamese network with a prediction encoder $\mathcal{P}$ and a target encoder $\mathcal{T}$~\cite{xu2021rethinking, chen2020simple}. The output of these networks are $\ell_2$-normalized: $p = \mathcal{P}(I) / 	\lVert \mathcal{P}(I) \rVert_2,$ and $z = \mathcal{T}(I') / 	\lVert \mathcal{T}(I') \rVert_2.$ 
Given a positive pair from a minibatch of size $N$, the other $2(N-1)$ examples are treated as negative examples. The objective then is to minimize the Info-NCE loss~\cite{oord2018representation}. When learning with negatives, $\mathcal{P}$ and $\mathcal{T}$ typically share the same architecture and model parameters.

\subsection{\model}
We propose a novel approach for learning representations by applying masking image modeling at the individual image level, paired with image-level similarity using either sampled frames or augmented images. Unlike previous methods that inefficiently replicate images to mimic video input, thereby utilizing more computational resources, our methodology treats short video segments as augmented instances of a single view. This perspective not only enhances the efficiency of the learned representations but also significantly broadens the applicability of our model. \model offers a versatile "plug and play" solution for image-based tasks. Furthermore, our model can easily be fine-tuned for video tasks and adapted to videos of varying sizes, unlike the traditional 16 frames. 
Figure~\ref{fig:vic-mae} shows an overview of our model. 

Given a video with $T$ frames $\{I_1, I_2, \cdots, I_T\}$, we sample two frames $I_i, I_j$ as a positive pair input during one training step. We augment single images when they appear in a batch. Notice that our model sees a batch comprising frames and images. After an input image tokenizer layer, we obtain a set of patch-level token representations of $X_i$ and $X_j$ for each frame. Then, we apply token masking by generating a different random mask $M_i$ and $M_j$ and apply them to both of the corresponding input frames to obtain a subset of input visible tokens $X_i^{(v)}$ and $X_j^{(v)}$. These visible token sets are then forwarded to a ViT encoder, which computes a set of representations $f_{\theta}(X_i^{(v)})$ and $f_{\theta}(X_j^{(v)})$ respectively. Finally, for the first image, we compute $\hat{I}_i = d_{\phi}(f_{\theta}(X_i^{(v)} + f_m))$ where we have added a mask token $f_m$ to let the decoder know which patches were masked and allows to predict patch-shaped outputs through $\hat{I}_i$. These output patches are then trained to minimize the $\ell_2$ loss with the true patches in the input image: 
\begin{equation}
    \mathcal{L}_i^{\text{MASK}} = \lVert \hat{I}_i - I_i \rVert_2^2.
\end{equation} 
To apply contrastive pre-training we use a separate prediction branch in the network by applying a global pooling operator $\Omega$ over the output representations $f_{\theta}(X_i^{(v)})$ from the main branch and $f_{\theta}(X_j^{(v)})$ from the siamese copy of the network. 
This step simplifies the formulation of our method and avoids using additional losses or the \texttt{gradient-stop} operator as in SimSiam~\cite{chen2021exploring} to avoid feature representation collapse since the pooled features can not default to the zero vector as they also are being trained to reconstruct patches. We experiment using various aggregation methods, including \textit{mean} pooling, \textit{max} pooling, and \textit{generalized mean}~(GeM) pooling \cite{radenovic2018fine}.

These global representations are then forwarded to a predictor encoder $\mathcal{P}$ and a target encoder $\mathcal{T}$ to obtain frame representations:
$$p_i \triangleq  \mathcal{P}(\Omega(f_{\theta}(X_i^{(v)}))) / 	\lVert \mathcal{P}(\Omega(f_{\theta}(X_i^{(v)})))) \rVert_2,$$
and 
$$z_j \triangleq  \mathcal{T}(\Omega(f_{\theta}(X_j^{(v)}))) / 	\lVert \mathcal{T}(\Omega(f_{\theta}(X_j^{(v)})))) \rVert_2$$
respectively.  The predictor network $\mathcal{P}$ and target network $\mathcal{T}$ are symmetrical and we use standard blocks designed for contrastive learning~\cite{bardes2021vicreg, chen2020simple, chen2021exploring}. These blocks consist of a Linear $\rightarrow$ BatchNorm1d $\rightarrow$ ReLU block repeated $2$ times. From these representations, we apply the InfoNCE contrastive learning loss as follows: 
\begin{equation}
    \begin{aligned}
    \mathcal{L}^{\text{NEG}}_{p_i,z_j} = -\log \frac{\text{exp}(p_i \cdot z_j / \tau)}{\sum^{2N}_{k=1} \mathbbm{1} [p_i \neq z_k] \text{exp}(p_i \cdot z_k / \tau)},
    \end{aligned}
    \label{eq: neg_loss}
\end{equation}
where the denominator includes a set of negative pairs with representations $z_k$ computed for frames from other videos, the same video but at a time longer than the selected time shift and images in the same batch, $\mathbbm{1} [p_i \neq z_k] \in \{0, 1\}$ is an indicator function evaluating to $1$ when $p_I \neq z_k$ and $\tau$ denotes a temperature parameter. 

The final loss is $\mathcal{L} = \mathcal{L}^{\text{MASK}} + \lambda \mathcal{L}^{\text{NEG}}$, where $\lambda$ is a hyperparameter controlling the relative influence of both losses. In practice, we use a schedule to gradually introduce the contrastive loss and let the model learn good local features at the beginning of training. 

\section{Experiment Settings}
\label{sec:experiments}

\newcommand{\grayfont}{\color{gray}}
\newcommand{\lightfont}{\footnotesize\color{black!60}}

\begin{table}[t!]
\small
\renewcommand{\arraystretch}{1.1}
\centering
\caption{\textbf{Transfer learning results from video and image pre-training to various datasets using the ViT/L-16 backbone}. The pre-training data is a video dataset (MiT, K600, K700, or K400) and/or image dataset (IN1K). All self-supervised methods are evaluated end-to-end with supervised finetuning on IN1K, Kinetics-400, Places365, and SSv2. Best results are in bold.  Results of MAE, ST-MAE, and VideoMAE for out-of-domain data were taken from Girdhar~\etal\cite{girdhar2023omnimae}.}
\label{tab:main_result}
\resizebox{\textwidth}{!}{%
\begin{tabular}{l l c c cc c cc}
\toprule
&\multirow{2}{*}{\textbf{Method}} & \multirow{2}{*}{\textbf{Arch.}} & \multirow{2}{*}{\textbf{Pre-training Data}} & \multicolumn{2}{c}{\textbf{In-Domain}} && \multicolumn{2}{c}{\textbf{Out-of-Domain}} \\ 
\cmidrule{5-6} \cmidrule{8-9}
& &  &  & IN1K & K400 && Places-365 & SSv2 \\ 
\midrule
\multirow{5}{*}{\rotatebox[origin=c]{90}{{\grayfont\footnotesize Supervised}}}&

\grayfont ViT~\cite{dosovitskiy2020image}
\emph{\lightfont ICML'20}
&\grayfont  ViT-B&\grayfont\footnotesize IN1K & \grayfont 82.3 & \grayfont 68.5 && \grayfont 57.0 & \grayfont 61.8 \\

&\grayfont ViT~\cite{dosovitskiy2020image}
\emph{\lightfont ICML'20}
&\grayfont ViT-L &\grayfont\footnotesize IN1K & \grayfont 82.6 & \grayfont 78.6 && \grayfont 58.9 & \grayfont 66.2 \\

&\grayfont COVeR \cite{zhang2021co}
\emph{\lightfont arXiv'21}
&\grayfont TimeSFormer-SR & \grayfont\footnotesize JFT-3B+ K400+ MiT + IN1K & \grayfont 86.6 & \grayfont 87.2 && \grayfont - & \grayfont 70.9 \\

&\grayfont OMNIVORE \cite{girdhar2022omnivore}
\emph{\lightfont CVPR'22}
&\grayfont ViT-B & \grayfont\footnotesize IN1K + K400 + SUN RGB-D & \grayfont 84.0 & \grayfont 83.3 && \grayfont 59.2 & \grayfont 68.3 \\

&\grayfont OMNIVORE \cite{girdhar2022omnivore}
\emph{\lightfont CVPR'22}
&\grayfont ViT-L & \grayfont\footnotesize IN1K + K400 + SUN RGB-D & \grayfont 86.0 & \grayfont 84.1 && \grayfont -- & \grayfont -- \\

&\grayfont TubeViT 
\cite{piergiovanni2022rethinking}
\emph{\lightfont CVPR'23}
&\grayfont ViT-B & \grayfont\footnotesize K400 + IN1K & \grayfont 81.4 & \grayfont 88.6 && \grayfont -- & \grayfont -- \\

&\grayfont TubeViT \cite{piergiovanni2022rethinking}
\emph{\lightfont CVPR'23}
&\grayfont ViT-L & \grayfont\footnotesize K400 + IN1K & \grayfont -- & \grayfont 90.2 && \grayfont -- & \grayfont 76.1 \\

\midrule

\multirow{14}{*}{\rotatebox[origin=c]{90}{{\footnotesize Self-Supervised}}}

&MAE~\cite{he2022masked} \emph{\lightfont CVPR'22}
& ViT-B &\footnotesize IN1K & {83.4} & -- && 57.9 & 59.6 \\

&MAE \cite{he2022masked}
\emph{\lightfont CVPR'22}
& ViT-L & \footnotesize IN1K & {85.5} & 82.3 && 59.4 & 57.7 \\

&ST-MAE~\cite{feichtenhofer2022masked}
\emph{\lightfont NeurIPS'22}
& ViT-B &\footnotesize K400 & 81.3 & 81.3 && 57.4 & 69.3 \\

&ST-MAE~\cite{feichtenhofer2022masked}
\emph{\lightfont NeurIPS'22}
& ViT-L &\footnotesize K400 & 81.7 & 84.8 && 58.1 & 73.2 \\

&VideoMAE \cite{tong22} 
\emph{\lightfont NeurIPS'22}
& ViT-B & \footnotesize K400 & 81.1 & 80.0 && -- & 69.6 \\

&VideoMAE \cite{tong22} 
\emph{\lightfont NeurIPS'22}
& ViT-L & \footnotesize K400 & -- & 85.2 && -- & 74.3 \\

&OmniMAE \cite{girdhar2023omnimae}
\emph{\lightfont CVPR'23}
& ViT-B & \footnotesize K400 + IN1K & 82.8 & 80.8 && 58.5 & 69.0 \\

&OmniMAE \cite{girdhar2023omnimae}
\emph{\lightfont CVPR'23}
& ViT-L & \footnotesize K400 + IN1K & 84.7 & 84.0 && 59.4 & 73.4 \\

\cmidrule{2-9}
&\model & ViT-L & \footnotesize K400 & 85.0 & 85.1 && 59.5 & 73.7 \\

&\model & ViT-L & \footnotesize MiT & 85.3 & 84.9 && 59.7 & 73.8 \\

&\model & ViT-B & \footnotesize K400 + IN1K & 83.0 & 80.8 && 58.6 & 69.5 \\

&\model & ViT-L & \footnotesize K400 + IN1K & 86.0 & 86.8 && 60.0 & 75.0 \\

\cmidrule{2-9}

&\model & ViT-B & \footnotesize  K710+ MiT + IN1K & 83.8 & 80.9 && 59.1 & 69.8 \\

&\model & ViT-L & \footnotesize  K710 + MiT + IN1K & \textbf{87.1} & \textbf{87.8} && \textbf{60.7} & \textbf{75.9}\\

\bottomrule
\end{tabular}
}
\end{table}

We perform experiments to demonstrate the fine-tuning performance of our method on ImageNet-1k and other image recognition datasets. We also evaluate our method on the Kinetics-400 dataset~\cite{kinetics-400} and Something Something-v2~\cite{goyal2017something} for action recognition to show that our model is able to maintain performance on video benchmarks. Full details are in the \supp.

\noindent
\textbf{Architecture}. We use the standard Vision Transformer (ViT) architecture~\cite{dosovitskiy2020image} and conduct experiments fairly across benchmarks and methods using the ViT-B/16 and ViT-L/16 configurations. For masked image modeling, we use a small decoder as proposed by He~\etal~\cite{he2022masked}. Finetunig on images requires no changes since this resembles the pre-training configuration. Finetuning on videos is as follows: we initialize the temporal tokenizer by replicating the spatial tokens along the temporal dimension scaled by the length of the video, similarly, we initialize the MHA parameters by replicating them but skip the scaling for them. We use the standard of finetuning on videos of 16 frames, skipping 4.

\noindent
\textbf{Pre-Training}. We adopt Moments in Time~\cite{momentsintime}, Kinetics-400 ~\cite{kinetics-400}, and ImageNet-1k~\cite{deng2009imagenet} as our main datasets for self supervised pre-training. They consist of $\sim$1000K and $\sim$300K videos of varied length respectively, and $\sim$1.2M images for Imagenet-1k. We sample frames from these videos using distant sampling, which consists of splitting the video into non-overlapping sections and sampling one frame from each section. Frames are resized to a 224 pixel size, horizontal flipping, and random cropping with a scale range of $[0.5, 1]$, as the only data augmentation transformations on video data. Random cropping (with flip and resize), color distortions, and Gaussian blurring are used for the image modality. For our largest training run, we combine the training sets of Kinetics-400 ~\cite{kinetics-400}, Kinetics-600~\cite{carreira2018short}, and Kinetics-700\cite{carreira2019short}, with duplicates removed based on YouTube IDs. We also exclude K400 videos used for evaluation from training to avoid leakage. This process results in a unique, diverse dataset of $\sim$665K samples, which we label K710, following \cite{wang2023videomae}.

\noindent
\textbf{Settings}. We follow previously used configurations for pre-training~\cite{he2022masked, feichtenhofer2022masked}. We use the AdamW optimizer with a batch size of 512 per device. We evaluate the pre-training quality by end-to-end finetuning. When evaluating on video datasets we follow the common practice of multi-view testing: taking $K$ temporal clips ($K=7$ on Kinetics) and for each clip taking 3 spatial views to cover the spatial axis (this is denoted as $K \times 3$). The final prediction is the average of all views.

\section{Results and Ablations}
\label{sec:results}
We first perform experiments to analyze the different elements of the \model framework. All the experiments are under the \textit{learning with negative pairs} setting using mean pooling over the ViT features. Linear evaluation and end-to-end finetuning runs are done over 100 epochs for ImageNet-1k, see \supp for more details. For our ablations, we restrict ourselves to the ViT-B/16 architecture pre-trained over 400 epochs unless specified otherwise.
% \vspace{-0.2in}
\subsection{Main result}
\begin{table}[t!]
\footnotesize
\centering
\setlength\tabcolsep{1.8pt}
\renewcommand{\arraystretch}{1.2}
\caption{\textbf{Comparison of transfer learning performance of our approach} with supervised baselines across 8 natural image classification datasets. All results correspond to linear evaluation. Best results are shown in bold. \textdaggerdbl MAE trained on MiT and K400 randomly sample a frame from the video to compute a reconstruction loss; these models are trained and evaluated by us. See \supp for more evaluation of transfer learning performance.}
\label{tab:model-performance}
\resizebox{\textwidth}{!}{%
    \begin{tabular}{llccccccccccccc}
    
    \toprule

    &\textbf{Model} & Pre-train. & Food & CIFAR10  & CIFAR100  & Birdsnap  & SUN397  & VOC2007  & DTD  & Caltech101  \\
    \midrule

\multirow{4}{*}{\rotatebox[origin=c]{90}{{\grayfont\footnotesize ViT/B-16}}}&
MAE \cite{he2022masked} \textdaggerdbl& K400 & 74.54 & 94.86 & 79.49 & 46.51 & 64.33 & 83.07 & 78.01  & 93.28 \\

&MAE \cite{he2022masked} \textdaggerdbl& MiT & 76.23 & 94.47 & 79.50 & 47.98 & 65.32 & 83.46 & 78.21 & 93.08 \\

&\model (ours) & K400 & 76.56 & 93.64 & 78.80 & 47.56 & 64.75 & 83.74 & 78.53 & 92.27 \\

&\model (ours) & MiT & \textbf{77.39} & \textbf{94.92} & \textbf{79.88} & \textbf{48.21} & \textbf{65.64} & \textbf{84.77} & \textbf{79.27} & \textbf{93.53} \\
\midrule
\multirow{4}{*}{\rotatebox[origin=c]{90}{{\grayfont\footnotesize ViT/L-16}}}&
MAE \cite{he2022masked} & IN1K & 77.5 & 95.0 & 82.9 & 49.8 & 63.2 & 83.3 & 74.5 & 94.8 \\

&OmniMAE \cite{girdhar2023omnimae} & SSv2+IN1K & 76.2 & 94.2 & 82.2 & 50.1 & 62.6 & 82.7 & 73.9 & 94.4 \\

&\model (ours) & IN1K+K400 & 81.9 & 95.6 & 85.4 & 52.8 & 67.3 & 84.2 & 76.8& 94.9 \\

&\model (ours) & K710+MiT+IN1K &  \textbf{82.9} & \textbf{96.8} & \textbf{86.5} & \textbf{53.5} & \textbf{68.1} & \textbf{85.3} & \textbf{77.8}  & \textbf{96.1} \\

    \bottomrule
    \end{tabular}
}
\vspace{-0.2in}
\end{table}

Our main result evaluates \model on two in-domain datasets that were used during training for most experiments: ImageNet-1K (images) and Kinetics-400 (video), and two out-of-domain datasets that no methods used during training: Places-365~\cite{zhou2017places} (images) and Something-something-v2 (video). Table~\ref{tab:main_result} shows our complete set of results including comparisons with the state-of-the-art on both supervised representation learning (typically using classification losses), and self-supervised representation learning (mostly using masked image modeling). We consider mostly recent methods building on visual transformers as the most recent TubeViT~\cite{piergiovanni2022rethinking} which relies on this type of architecture.\footnote{Previous methods also use different backbones~\cite{gordon2020watching,xu2021rethinking,wu2021contrastive}~\ie ResNet-50. They obtain 54.5\%, 33.8\%, and 55.6\% top-1 accuracies on linear evaluation on ImageNet-1k. Since those works do not use the same setting, we do not include them here.}

Our most advanced version of \model trained on five datasets (Kinetics-400, Kinetics-600, Kinetics-700, Moments in Time, and Imagenet-1k) using the ViT-Large architecture performs the best across all metrics on all datasets compared to all previous self-supervised representation learning methods and even outperforms the supervised base model OMNIVORE~\cite{girdhar2022omnivore} on Imagenet-1k with a top-1 accuracy of 87.1\% vs 86\%. As well as, COVeR \cite{zhang2021co} a model trained on a similar data mix, except that it uses more images, COVeR gets 86.6\% vs 87.1\% on Imagenet-1k.
\model also comes a close second to other supervised methods and roughly matches the performance of TubeViT~\cite{piergiovanni2022rethinking} which obtains 76.1\% top-1 accuracy on Something something-v2 compared to our 75.9\% top-1 accuracy. When compared to the current self-supervised state-of-the-art OmniMAE using the same ViT-Large architecture and the same datasets for pre-training (Kinetics-400 and Imagenet-1k), \model also outperforms OmniMAE in all benchmarks (Imagenet: 86\% \vs 84.7\%, Kinetics-400: 86.8\% \vs 84\%, Places-365: 60\% \vs 59.4\% and SSv2: 75\% vs 73.4\%). 

Another important result is {\it video-to-image transfer}, where the model is only trained on video but its performance is tested on downstream image tasks. Table~\ref{tab:main_result} shows that when \model is trained on the Moments in Time dataset~\cite{momentsintime}, it achieves the best top-1 accuracy of 85.3\% for any self-supervised backbone model trained only on video. These results highlight the closing gap in building robust representations that can work seamlessly across image and video tasks. 

\subsection{Comparison with other contrastive masked autoencoders.}
Combining MAE with joint-embedding methods is non-trivial. In our first attempts, we used the [CLS] token as the representation and applied negative free methods such as VicReg~\cite{bardes2021vicreg}, and SimSiam~\cite{chen2021exploring} with limited success (See \supp). When combined with contrastive methods, we found it best to use a pooling operation over the ViT features similar to CAN~\cite{mishra2022simple}, as we find worse performance when the [CLS] token is used, like in C-MAE~\cite{huang2022contrastive}. The original MAE~\cite{he2022masked} is known to have poor linear evaluation performance, obtaining 68\% in IN1K linear evaluation when pre-trained on IN1K~\cite{he2022masked, lehner2023contrastive}. On the contrary, \mbox{SimCLR}~\cite{chen2020big} a model trained only using contrastive learning on IN1K achieves 73.5\%. Several works have tried to address this by combining contrastive learning with masked image modeling to get the best of both worlds. CAN~\cite{mishra2022simple}, C-MAE~\cite{huang2022contrastive} and \mbox{MAE-CT}~\cite{lehner2023contrastive} obtain linear evaluation accuracies of 74.0\%, 73.9, 73.4\%, respectively when trained on IN1K while \model obtains 74.0\% trained only on IN1K using ViT/B-16 pre-trained for 800 epochs to make the comparison fair. When using the K400 and IN1K datasets together for pre-training, we get 73.6\%, but we highlight that \model can now maintain good performance in videos and images using the same pre-trained model.

\begin{table}[t!]
    \centering
    \caption{\modelbold \textbf{ablation  experiments} with ViT/B-16. We present linear evaluation results on the ImageNet-1K dataset.}
    \vspace{-0.2in}
    \begin{subtable}[t]{.31\textwidth}
        \centering
        \caption{\textbf{Ablation on frame separation}.  $0$: sample same frame, $\text{D}$: distant sampling, and $>0$ continuous sampling.}
        \label{tab:ablation_frame}
        \resizebox{\linewidth}{!}{%
        \begin{tabular}{ccc}
            \toprule
            \textbf{Frame  separation} & \multicolumn{2}{c}{\textbf{ImageNet-1K}} \\
            \cmidrule{2-3}
             & Top-1 & Top-5 \\
            \midrule
            0 & 63.25 & 83.34 \\
            2 & 64.47 & 84.31 \\
            4 & 65.25 & 84.64 \\
            8 & 65.89 & 84.91 \\
            \midrule
            D & \textbf{67.66} & \textbf{86.22} \\
            \bottomrule
        \end{tabular}
        }
    \end{subtable}
    \hfill % This will add space between the two tables if necessary
    \begin{subtable}[t]{.31\textwidth}
        \centering
        \caption{\textbf{Ablation on pooling type}. The hyperparameter $\lambda$ is set to $0.025$ and introduced using a schedule.}
        \label{tab:ablation_loss}
        \resizebox{\linewidth}{!}{%
        \begin{tabular}{ccc}
            \toprule
            \textbf{Pooling type} & Top-1 & Top-5 \\
            \midrule
            GeM & 66.92 & 85.50 \\
            max & 67.01 & 85.59 \\
            mean & \textbf{67.66} & \textbf{86.22} \\
            \bottomrule
        \end{tabular}
        }
    \end{subtable}
    \hfill % This will add space between the two tables if necessary
    \begin{subtable}[t]{.31\textwidth}
        \centering
        \caption{\textbf{Ablation on different augmentations}.We use a combination of different color and spatial augs.}
        \label{tab:ablation_aug}
        \resizebox{\linewidth}{!}{%
        \begin{tabular}{cccc}
        \toprule
        \multirow{2}{*}{\begin{tabular}[c]{@{}c@{}}\textbf{Color}\\ \textbf{Augm}.\end{tabular}} & \multirow{2}{*}{\begin{tabular}[c]{@{}c@{}}\textbf{Spatial}\\ \textbf{Augm.}\end{tabular}} & \multicolumn{2}{c}{\textbf{ImageNet-1K}} \\
        \cmidrule{3-4}
         &  & Top-1 & Top-5 \\ 
         \midrule
        \checkmark &  & 65.40 & 84.03 \\
        & \checkmark & 66.03 & 85.01 \\
        \checkmark & \checkmark & \textbf{67.66} & \textbf{86.22}\\
        \bottomrule
        \end{tabular}
        }
    \end{subtable}
    \vspace{-0.2in}
\end{table}

\noindent
\subsection{Transfer Learning  Experiments}
In this section, we evaluate our pre-trained models from Table~\ref{tab:main_result} for transfer learning on downstream tasks.
\subsubsection{Video-to-image transfer learning performance.}
We evaluate transfer learning performance of \model across a diverse array of 12 downstream image classification tasks~\cite{bossard2014food, krizhevsky2009learning, berg2014birdsnap, xiao2010sun, krause20133d, maji2013fine, cimpoi2014describing, parkhi2012cats, fei2004learning, nilsback2008automated}. (Due to space constraints, we have shown the six most significant ones. See \supp for the full table.) Table~\ref{tab:model-performance} shows the results of four models based on a ViT/B backbone. We perform linear evaluation. We train two models using two video datasets. The first model is a baseline MAE model pre-trained on randomly sampled frames from videos on the Moments in Time and Kinetics-400 datasets. The second model is our full \model model pre-trained on each of the same two datasets. Our model significantly outperforms the other baselines on 9 out of 12 datasets, whereas the MAE trained on Kinetics is superior on only 3 (i.e. Cars, Aircraft, and Pets). When scaling the size of our models, we see that \model surpasses all models, including OmniMAE~\cite{girdhar2023omnimae} trained on SSv2+IN1K\footnote{These are the only publicly available checkpoints of OmniMAE}

\begin{table}[t!]
\footnotesize
\centering
\renewcommand{\arraystretch}{1.1}
\caption{\textbf{COCO object detection and segmentation} using a ViT-B Mask R-CNN baseline. All entries use data without labels.}
\label{tab:detection-results}
\begin{tabular}{@{}lccc@{}}
\toprule
\textbf{Method} & \textbf{pre-train data} & \textbf{AP$_{Box}$} & \textbf{AP$_{Mask}$} \\ \midrule
MAE~\cite{he2022masked} & IN1K & 50.3 & 44.9 \\
C-MAE~\cite{huang2022contrastive} & IN1K & 52.4 & 46.5 \\
\model & IN1K+K400 & 52.5 & 46.5 \\
\model & IN1K+K710+MiT & {\bf 53.2} & {\bf 46.9} \\ \bottomrule
\end{tabular}
\end{table}

\subsubsection{Object detection and segmentation.}
We finetune Mask R-CNN~\cite{he2017mask} end-to-end on the COCO dataset. We adapted the ViT backbone to be used with the FPN, following the recipe outlined in Li~\etal~\cite{li2022exploring}. We apply this approach to \model and take the other results from their respective paper. See Table~\ref{tab:detection-results}. 
\noindent
Compared with previous methods, \model outperforms other approaches under the same configurations. Specifically, when utilizing the combined IN1K+K400 dataset, \model achieves a box AP of 52.5 and a mask AP of 46.5, slightly improving over C-MAE, which stands at 52.4 for box AP and 46.5 for mask AP. More notably, with the expanded dataset of IN1K+K710+MiT, \model significantly advances the state-of-the-art, achieving the highest reported scores of 53.2 for box AP and 46.9 for mask AP. 
\vspace{-0.2in}
\subsection{Ablations}
We investigate the effect of scaling the data used to train \model, the effect of the ratio of image to videos in pre-training, our choice of frame separation, the choice of pooling operator, and the choice of data augmentations. An extra ablation probing the temporal representation learning of our method can be found in the \supp.

\begin{figure}[t!]
    \centering
    \begin{subfigure}{.48\textwidth}
        \centering
        \vspace{0pt} % Ensures alignment at the top
        \includegraphics[width=\linewidth]{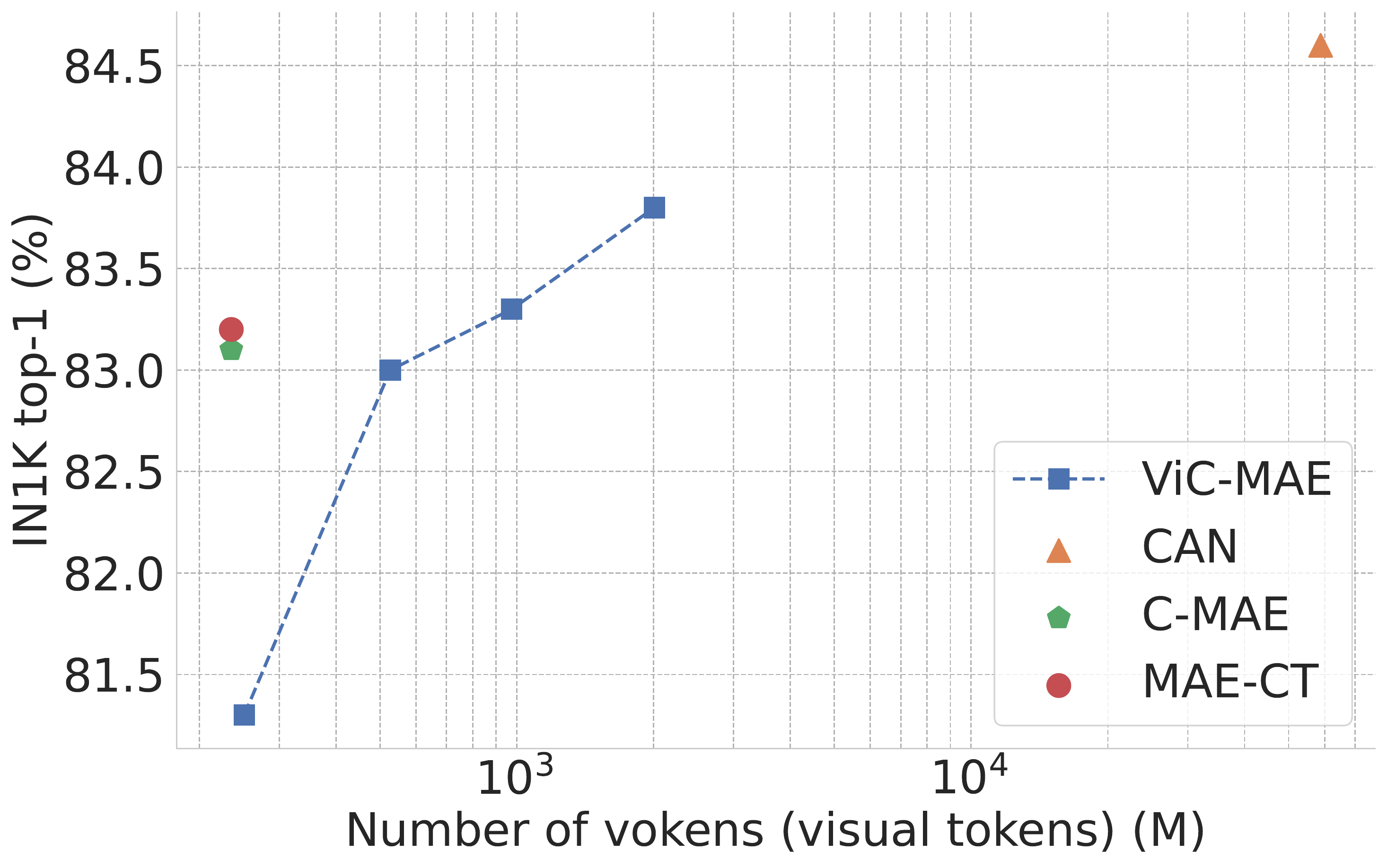} % Adjust the size as needed
        \caption{\model ViT/B-16 finetuned on IN1K for 100 epochs, compared with CAN pre-trained on JFT-300M, C-MAE, and MAE-CT pre-trained on ImageNet-1K. We increase the amount of data points by adding more video datasets. We can see that our model reaches similar accuracy with $\approx4.25$M data points compared to the $300$M of the JFT-300M dataset.}
        \label{fig:scaling}
    \end{subfigure}\hfill %
    %\vskip 0pt
    \begin{subfigure}{.48\textwidth}
        \centering
        \vskip 0pt
        \vspace{0pt} % Ensures alignment at the top
        \includegraphics[width=\linewidth]{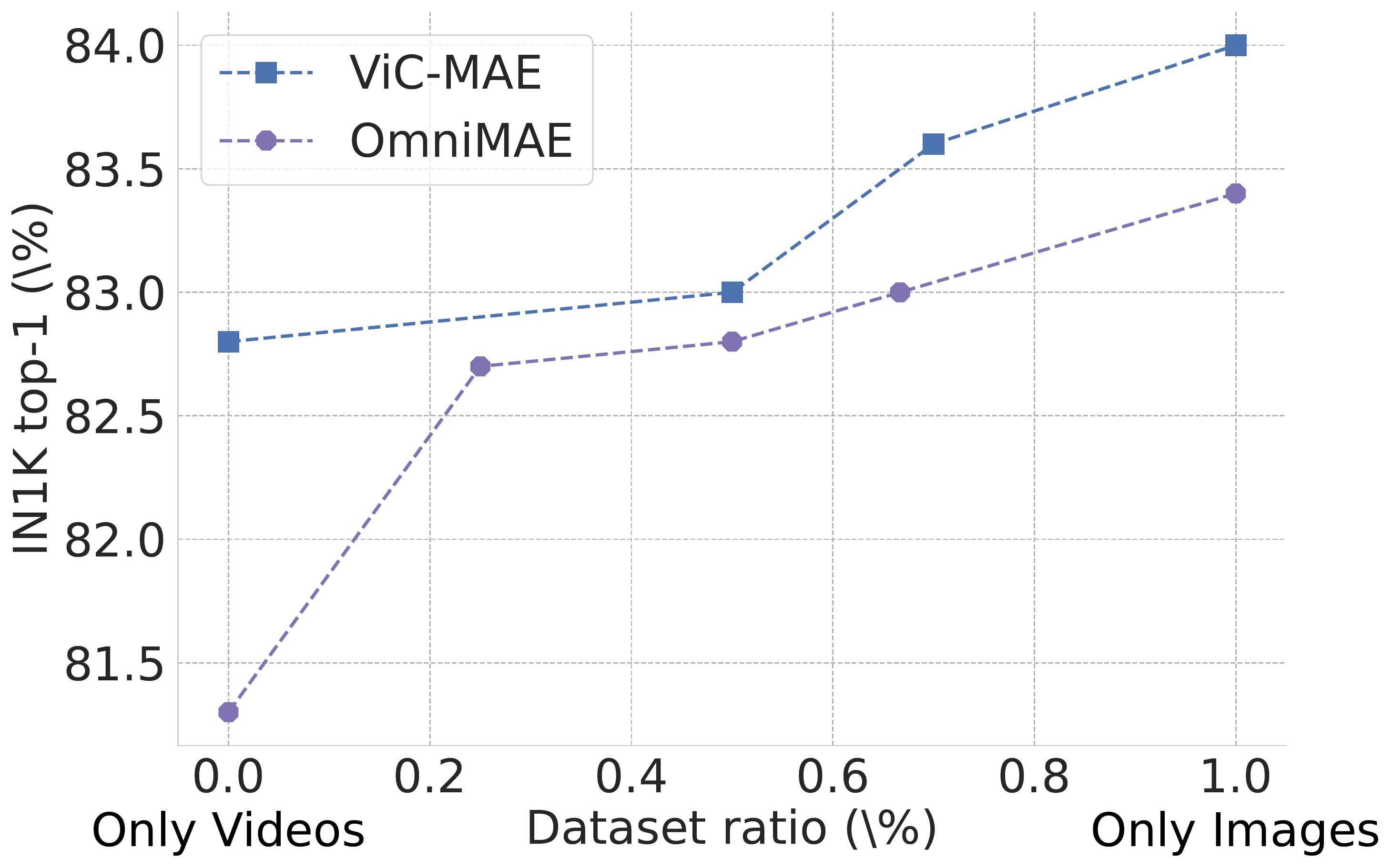} % Adjust the size as needed
        \caption{\model using the ViT/B-16 architecture finetuned on IN1K for 100 epochs, compared with OmniMAE. We vary the ratio of images vs video in the dataset, from no images to only images. We can see that \model can better utilize the videos and images on the dataset compared to masking-only pre-training.}
        \label{fig:dataset_ratio}
    \end{subfigure}
    % Optional: Common caption for both figures
    \caption{Additional comparisons with the state-of-the-art and recently proposed methods.}
    \label{fig:combined}
    \vspace{-0.2in}
\end{figure}

\noindent
\textbf{Influence of pre-training data.} We perform an ablation study to the effect of scaling the data points seen by the model. The pre-training data includes Kinectis-400, ImageNet-1K, Kinectis-600 + Kinectis-700, and the Moments in Time datasets added in that order. We pre-train a ViT/B-16 using \model for 400 epochs. As illustrated in Figure~\ref{fig:scaling}, as we progressively increase the dataset size, our \model, shows a steady increase in IN1K top-1 accuracy. This is remarkable when compared to CAN~\cite{mishra2022simple}, pre-trained on the JFT-300M dataset for 800 epochs that only reaches an accuracy of 84.4\%. This shows that our model, when supplied with only about 1.5\% of the data that CAN was trained on (4.25M vs. 300M), can achieve comparable accuracy levels.

\noindent
\textbf{Contrastive vs Masking-only pre-training} We perform an ablation study to the effect of varying the ratio of images to video in the dataset by replicating the entire dataset; notice that the number of training updates changes when doing this. The pre-training data includes Kinectis-400 and ImageNet-1K. We pre-train a ViT/B-16 using \model for 400 epochs. As illustrated in Figure~\ref{fig:dataset_ratio}, as we progressively increase the ratio of images to videos, our \model, surpasses the OmniMAE model~\cite{xu2021rethinking}, meaning that contrastive plus masking pre-training is better able to use image and video data than masking-only pre-training.

\noindent
\textbf{Frame separation}.  We aim to explore the effect of frame separation on model performance. We follow the two methods of sampling frames from Xu~et.al~\cite{xu2021rethinking}. Results are shown in Table \ref{tab:ablation_frame}.
The first approach, \textit{Continuous sampling}, involves selecting a start index $i$ and sampling a frame within $(i, i+\delta]$, where $\delta$ represents the frame separation, with a separation of $0$ meaning identical frames for predictor and target networks. The second, \textit{Distant sampling}, divides the video into $n$ equal intervals, corresponding to the number of frames for contrastive learning, and randomly selects one frame from each interval.
 
In our experiment, we observe that increasing the frame separation when using \textit{continuous sampling} increases model performance. We observe the best performance using \textit{distant sampling} with $n=2$ (labeled $D$ in Table \ref{tab:ablation_frame}). We posit that further increasing frame separation offers potentially stronger augmentations. In the following experiments, we only use strong spatial augmentations combined with distant frame sampling. 

\noindent
\textbf{Pooling type}. We test which operator $\Omega$ used to aggregate local features performs best at producing global features. We report our results in Table~\ref{tab:ablation_loss}. We try common types of pooling (\textit{mean, max}) as well as, \textit{generalized mean pooling}. 
We found \textit{mean} to be more effective in creating a global representation for video, and we use it for all other experiments.

\noindent
\textbf{Adding strong augmentations to video frames}
In our ablation study, we investigated the necessity of strong color augmentations for video frames during joint training with the target encoder, as commonly applied to images. The findings, detailed in Table~\ref{tab:ablation_aug}, indicate a performance decrease of over 2\% in linear evaluation on the Imagenet dataset when applying solely color augmentations without spatial adjustments. Interestingly, employing color and spatial augmentations does not outperform strong spatial augmentations alone. This diverges from prior approaches that rely heavily on color augmentations for effective contrastive learning, suggesting that the inherent temporal variations in video frames may suffice. However, for image datasets, the combination of strong color and spatial augmentations remains necessary.

\subsection{Limitations}
Our proposed model is able to learn representations from video and image data that transfer to several downstream tasks and surpasses previous models on the same set-up. 
Given a similar setup, \model matches prior results on Kinetics-400, trailing only to supervised models such as TubeViT~\cite{piergiovanni2022rethinking} by $7.1\%$ on ViT/B-16 and $2.4\%$ on ViT/L-16, and MVT~\cite{yan2022multiview} slightly on ViT/B-16 but surpasses it by $3.5\%$ on ViT/L-16. It also exceeds MViTv1~\cite{fan2021multiscale}, TimeSformer~\cite{bertasius2021space}, and ViViT~\cite{arnab2021vivit} by margins up to $7.3\%$ on ViT/L-16. Compared to self-supervised models, \model falls behind MaskFeat~\cite{wei2022masked} by $0.7\%$ on ViT/B-16 but excels on ViT/L-16 by $3.5\%$. Our model surpasses V-JEPA~\cite{bardes2023v} by margins up to $1.1\%$ but falls behind VideMAEV2~\cite{wang2023videomae} by $0.8\%$ on ViT/L-16.
It is slightly outperformed by DINO~\cite{caron2021emerging} and more substantially by models using extra text data or larger image datasets, such as UMT~\cite{li2023unmasked}, MVD~\cite{wang2023masked}, and UniFormerV2~\cite{li2022uniformerv2}, by up to $4.2\%$ on ViT/B-16 and $2.8\%$ on ViT/L-16. Future work could consider leveraging additional weak supervision through other modalities such as text~\cite{shrivastava2023clip,he2024improved,Cai_2024_CVPR}, audio~\cite{singh2024looking,tang2024avicunaaudiovisualllminterleaver}, 3D geometry~\cite{sriram2023home} or automatically generated data~\cite{synclip_2023_ICCV,he2024learningmodelsdatavisual}.

When compared against state-of-the-art ImageNet-pretrained models with comparable computational resources, video-based models, including ours, typically fall short. However, including image and video modalities shows promise in boosting performance. Against models using masked image modeling and contrastive learning, \model modestly surpasses MAE~\cite{he2022masked} by $1.6\%$, MaskFeat~\cite{wei2022masked} by $1.4\%$ and iBOT~\cite{zhou2021ibot} by $0.5\%$ with the ViT/L-16 architecture. It also edges out MoCov3~\cite{chen2020improved} and BeiT~\cite{bao2021beit} by $3\%$ and $1.9\%$ respectively on the same architecture. Yet, it lags behind DINOv2 \cite{oquab2023dinov2} by $1.2\%$ for ViT/L-16. When compared to supervised models using additional image data, such as DeiT-III~\cite{touvron2022deit} and SwinV2~\cite{liu2022swin} and the distilled ViTs' from \cite{dehghani2023scaling}, our model shows a lag behind of $0.8\%$, $1.2\%$ and $2.5\%$ respectively on ViT/L-16. These results show that the gap from models pre-trained purely on video still exists, but we believe \model pre-trained on image and video data is a step forward in closing that gap.

\vspace{-0.1in}
\section{Conclusion}
\label{sec:conclusion}
In this work, we introduce \model, a method that allows to use unlabeled videos and images to learn useful representation for image recognition tasks. We achieve this by randomly sampling frames from a video or creating two augmented views of an image and using contrastive learning to pull together inputs from the same video and push apart inputs from different videos, likewise, we also use masked image modeling on each input to learn good local features of the scene presented in each input. The main contribution of our work is showing that it is possible to combine masked image modeling and contrastive learning by pooling the local representations of the MAE prediction heads into a global representation used for contrastive learning.
The design choices that we have taken when designing \model show that our work is easily extensible in various ways. For example, improvements in contrastive learning for images can be directly adapted into our framework. Likewise, pixel reconstruction can be replaced by features important for video representation such as object correspondences or optical flow.

\section*{Acknowledgements}
The authors would like to thank Google Cloud and the CURe program from Google Research for partially providing funding for this research effort. We are also thankful for support from the Department of Computer Science at Rice University, the National Science Foundation through NSF CAREER Award \#2201710, and the Ken Kennedy Institute at Rice University. We also thank anonymous reviewers for their feedback and encouragement.

\bibliographystyle{splncs04}
\bibliography{references}

{
\clearpage 
\appendix
\appendix
\section{Implementation Details}\label{supp:impl}
\paragraph{\model architecture.}
\begin{wrapfigure}{r}{0.61\textwidth}
\vspace{-0.25in}
\begin{algorithm}[H]
  \caption{\model \ PyTorch pseudocode.}
  \label{alg:method}
    \definecolor{codeblue}{rgb}{0.25,0.5,0.5}
    \definecolor{codekw}{rgb}{0.85, 0.18, 0.50}
    \newcommand{\algofontsize}{7pt}
    \lstset{
      backgroundcolor=\color{white},
      basicstyle=\fontsize{\algofontsize}{\algofontsize}\ttfamily\selectfont,
      columns=fullflexible,
      breaklines=true,
      captionpos=b,
      commentstyle=\fontsize{\algofontsize}{\algofontsize}\color{codeblue},
      keywordstyle=\fontsize{\algofontsize}{\algofontsize}\color{codekw},
    }
\begin{lstlisting}[language=python]
# V[N, T, C, H, W] - minibatch (T=1 for images)
# tau: temperature coefficient
# clambda: contrastive coefficient

for V in loader:
    # Distant sampling
    f_i, f_j = random_sampling(V) 
    # Patch embeddings and position encodings
    x_i = patch_embedd(f)
    x_i += pos_embedd
    # Mask out patches
    x_i, mask_i, ids_restore_i = random_masking(x)
    # Patchify, add pos_embed and mask out f_i ...
    # Forward frames on masked input
    x_i = frame_encoder(x_i) # [N, L_msk, D]
    x_j = frame_encoder(x_j) # [N, L_msk, D]
    # Pool features
    x_pool_i = pooling(x_i) # [N, D]
    x_pool_j = pooling(x_j) # [N, D]
    # Project and normalize
    p_i = l2_normalize(projector(x_pool_i), dim=1)
    z_j = l2_normalize(projector(x_pool_j), dim=1)
    # Predict pixels
    pred = decoder(x_i) # after adding mask tokens
    
    # compute pixel loss
    target = patchify(f_i)
    loss_pixel = (pred - target) ** 2
    loss_pixel = loss_pixel.mean(dim=-1)  # [N, L]
    loss_pixel = (loss * mask).sum() / mask.sum() 
    # compute contrastive loss
    loss_cons =  ctr(p_i, z_j) + ctr(z_j, p_i) 
    # compute final loss
    loss = loss_pixel + clambda * loss_cons


def ctr(p, z):
    # similarity matrix [N, N]
    sim = einsum('nl,nl->nn', p, z) * exp(tau)
    # compute info-nce loss
    labels = range(N) # positives are in diagonal
    loss = cross_entropy_loss(sim, labels)
    return 2 * loss
\end{lstlisting}
\end{algorithm}
\vspace{-0.3in}
\end{wrapfigure}

We will release code and model checkpoints, along with the specific training configurations. We followed previous training configurations that also worked well for our models~\cite{he2022masked, feichtenhofer2022masked}. The pseudocode of \model is also provided in Algorithm~\ref{alg:method}.

We follow the standard ViT architecture~\cite{dosovitskiy2020image}, which has a stack of Transformer blocks, each of which consists of a multi-head attention block and an Multi-Layer Perceptron (MLP) block, with Layer Normalization (LN). A linear projection layer is used after the encoder to match the width of the decoder. We use sine-cosine position embeddings for both the encoder and decoder. For the projection and target networks, we do average pooling on the encoder features and follow the architecture of Bardes~\etal~\cite{bardes2021vicreg}, which consists of a linear layer projecting the features up to twice the size of the encoder and two blocks of linear layers that preserve the size of the features, followed by batch normalization and a ReLU non-linearity.

We extract features from the encoder output for fine-tuning and linear probing. We use the class token from the original ViT architecture, but notice that similar results are obtained without it (using average pooling).
\vspace{-5pt}
\paragraph{Video Loading.}
In order to prevent video loading from being a bottleneck on performance due to time spent on video decoding, we leverage the {\tt ffcv} library \cite{leclerc2023ffcv}, which we modify to support videos as a list of images in the WebP format. This allows us to significantly surpass the default PyTorch data loaders which can only read data in a synchronous fashion, resulting in the process being blocked until video decoding is complete. The use of {\tt ffcv} allows to perform training without the need of sample repetition as done in OmniMAE~\cite{girdhar2023omnimae} and ST-MAE~\cite{feichtenhofer2022masked} at the cost of a significantly larger storage requirement. We will also release the code for {\tt ffcv} to support videos.
\vspace{-5pt}
\paragraph{Pre-training.}
The default settings can be found in Table~\ref{tab:impl_vicmae_pretrain}. We do not perform any color augmentation, path dropping or gradient clipping. We initialize our transformer layer using xavier\_uniform \cite{glorot2010understanding}, as it is standard for Transformer architectures. We use the linear learning rate (\textit{lr}) scaling rule so that \textit{lr} = \textit{base\_lr}$\times$batchsize / 256~\cite{goyal2017accurate}.
\vspace{-5pt}
\paragraph{End-to-end finetuning.}
We follow common practice for end-to-end finetuning. Default settings can be found in Table~\ref{tab:impl_vicmae_finetune}. Similar to previous work, we use layer-wise \textit{lr} decay~\cite{he2022masked}.
\vspace{-5pt}
\paragraph{Linear evaluation.}
We follow previous work for linear evaluation results~\cite{he2022masked, feichtenhofer2022masked}. As previous work has found we do not use common regularization techniques such as mixup, cutmix, and drop path, and likewise, we set the weight decay to zero~\cite{chen2021empirical}. We add an extra batch normalization layer without the affine transformation after the encoder features. Default settings can be found in Table~\ref{tab:impl_vicmae_linear}.
\vspace{-5pt}

\begin{table}[ht]
\small
\centering
\caption{\textbf{Pre-training setting.}}
\label{tab:impl_vicmae_pretrain} 
% \vspace{-.5em}
\begin{tabular}{lc}
\toprule
config & value \\
\midrule
optimizer & AdamW \cite{loshchilov2017decoupled} \\
base learning rate & 1.5e-4 \\
weight decay & 0.05 \\
optimizer momentum & $\beta_1, \beta_2{=}0.9, 0.95$ \cite{chen2020generative} \\
batch size & 4096 \\
learning rate schedule & cosine decay \cite{loshchilov2016sgdr} \\
warmup epochs \cite{goyal2017accurate} & 40 \\
epochs & 800 \\
augmentation & hflip, crop [0.5, 1] \\
contrastive loss weight $\lambda$ & 0.025 \\
contrastive loss schedule & 0 until epoch 200 then 0.025 \\
\bottomrule
\end{tabular}
\vspace{-2em}

\end{table}

\begin{table}[ht]
\small
\centering
\caption{\textbf{End-to-end fine-tuning setting.}}
\label{tab:impl_vicmae_finetune}
\begin{tabular}{lcc}
\toprule
config & ViT/B & ViT/L \\
\midrule
optimizer & \multicolumn{2}{c}{AdamW} \\
optimizer momentum & \multicolumn{2}{c}{$\beta_1, \beta_2{=}0.9, 0.999$} \\
base learning rate  \\
\quad {IN1K} & 3e-3 & 0.5e-3 \\
\quad {P65} & \multicolumn{2}{c}{2e-3} \\
\quad {K400} & 1.6e-3 & 4.8e-3 \\
\quad {SSv2} & \multicolumn{2}{c}{1e-3} \\
weight decay & \multicolumn{2}{c}{0.05} \\
learning rate schedule &  \multicolumn{2}{c}{cosine decay} \\
warmup epochs & \multicolumn{2}{c}{5} \\
layer-wise lr decay \cite{clark2020electra,bao2021beit} & 0.65 & 0.75 \\
batch size & 1024 & 768 \\
training epochs \\
\quad {IN1K} & 100 & 50 \\
\quad {P65} & 60 & 50 \\
\quad {K400} & 150 & 100 \\
\quad {SSv2} & \multicolumn{2}{c}{40} \\
augmentation & \multicolumn{2}{c}{RandAug (9, 0.5) \cite{cubuk2020randaugment}} \\
label smoothing \cite{szegedy2016rethinking} & \multicolumn{2}{c}{0.1} \\
mixup \cite{zhang2017mixup} & \multicolumn{2}{c}{0.8} \\
drop path \cite{huang2016deep} & 0.1 & 0.2\\
\bottomrule
\end{tabular}

\end{table}

\begin{table}[t]
\small
\centering
\caption{\textbf{Linear evaluation setting.}
\label{tab:impl_vicmae_linear}}
\begin{tabular}{lc}
\toprule
config & value \\
\midrule
optimizer & SGD \\
base learning rate & 0.1 \\
weight decay & 0 \\
optimizer momentum & 0.9 \\
batch size & 4096 \\
learning rate schedule & cosine decay \\
warmup epochs & 10 \\
training epochs & 90 \\
augmentation & RandomResizedCrop \\
\bottomrule
\end{tabular}

\end{table}

\section{Combining MAE with Negative-Free Methods.}
\label{supp:tried_not_work}

\begin{table}[t!]
\centering
\setlength\tabcolsep{10pt}
\renewcommand{\arraystretch}{1.2}
\caption{\textbf{Combining MAE and contrastive methods is not trivial.} Linear evaluation on
the ImageNet-1K dataset using types of contrastive learning. We use the $[\text{CLS}]$ token as the global video representation and apply common contrastive methods, but these do not result on the best performance, which is obtained with our method.}
\label{tab:ours_vs_contrastive_loss}
\begin{tabular}{lcc}
 \toprule
 \multirow{2}{*}{\textbf{Method}} & \multicolumn{2}{c}{\textbf{ImageNet-1K}} \\
\cmidrule{2-3}
& Top-1 & Top-5 \\ 
\midrule
MAE \cite{he2022masked} + SiamSiam \cite{chen2021exploring} & 58.58 & 82.88 \\
MAE \cite{he2022masked} + VicReg \cite{bardes2021vicreg} & 63.86 & 84.07 \\
\model (ours) & \textbf{67.66} & \textbf{86.22}\\
\bottomrule
\end{tabular}

\end{table}

We tried to combine MAE with instance discrimination learning methods by using the $[\text{CLS}]$ token of the transformer as a global video feature representation. This representation allows us to use any instance discrimination learning loss without modifications to the underlying ViT transformer encoder. This combination works as follows: Sample two images  from a video or an image and its transformed version $I_i, I_j$ and perform patch-level masking. The two inputs are processed by the ViT model $f_{\theta}$ producing token representations $f_{\theta}(I_i) = \{x_i^{\text{CLS}}, x_i^1, x_i^2, \cdots, x_i^L\}$, where $L$ is the sequence length of the transformer model. This is divided into two disjoint sets. The set $\{x_i^1, x_i^2, \cdots, x_i^L\}$ represents the local features of the input $i$ and are used for masked image modeling following Eq. 1. Then, the $x_i^{\text{CLS}}$ token can be used as a global representation with a contrastive loss. 

We experiment with this approach using the SimSiam loss~\cite{chen2021exploring} and the VicReg loss~\cite{bardes2021vicreg}. We review here these methods and how to combine them with MAEs, but the reader is referred to the original works for a more in-depth explanation of these methods~\cite{chen2021exploring, bardes2021vicreg}.

\paragraph{SimSiam.}
A combination of SimSiam and MAE, which we refer to as {\em MAE + SimSiam} uses the  $x_i^{\text{CLS}}$ token which represents the global video representation as follows: We pass $x_i^{\text{CLS}}$ to a projector network $\mathcal{P}$ to obtain $p_i \triangleq  \mathcal{P}(x_i^{\text{CLS}}) / 	\lVert \mathcal{P}(x_i^{\text{CLS}}) \rVert_2$. A similar procedure is followed for input $j$, but the global representation is not passed to the projector network $\mathcal{P}$ in order to obtain $z_j \triangleq  x_i^{\text{CLS}} / 	\lVert x_i^{\text{CLS}} \rVert_2$. The SimSiam objective is then applied as follows:  
\begin{equation}
    \begin{aligned}
    \mathcal{L}^{\text{ \tiny SimSiam}}_{p_i,z_j} = \lVert p_i - z_j \rVert^2_2 = 2 (1 - p_i \cdot z_j).
    \end{aligned}
    \label{eq: pos_loss_siamsiam}
\end{equation}

\paragraph{VicReg.}
A combination of VicReg and MAE, which we refer to as {\em MAE + VicReg} uses the $x_i^{\text{CLS}}$ token which represents the global video representation as follows: We pass it to a projector network $\mathcal{P}$ to obtain $p_i \triangleq  \mathcal{P}(x_i^{\text{CLS}}) / 	\lVert \mathcal{P}(x_i^{\text{CLS}}) \rVert_2$, we repeat this procedure for input $j$ using the target network $\mathcal{T}$ to obtain $z_i \triangleq  \mathcal{T}(x_i^{\text{CLS}}) / 	\lVert \mathcal{T}(x_i^{\text{CLS}}) \rVert_2$. The loss is calculated at the embedding level on $p_i$ and $z_j$. The inputs are processed in batches, let us denote $P = [p^1, \cdots, p^n]$ and $Z = [z^1, \cdots, z^n]$, where each $p^m$ and $z^m$ are the global representation of video $m$ after the projector network and target network respectively in a batch of size $n$ vectors of dimension $d$. Let us denote by $p_l$ the vector composed of each value at dimension $l$ in all vectors in $P$. The variance loss of VicReg is then calculated as follows:
\begin{equation}
    \begin{aligned}
    v(P) = \frac{1}{d} \sum_{l=1}^d \text{max}(0, \gamma - S(p_i, \epsilon)),
    \end{aligned}
    \label{eq: var_loss_vicreg}
\end{equation}
where $S(z, \epsilon) = \sqrt{\text{Var}(z) + \epsilon}$ and $\gamma$ is a constant target value for the standard deviation, fixed to 1. The covariance loss of VicReg can be calculated as:
\begin{equation}
    \begin{aligned}
    c(P) = \frac{1}{d} \sum_{l\neq k}^d [\text{Cov}(p^m)]^2_{l,k},
    \end{aligned}
    \label{eq: cov_loss_vicreg}
\end{equation}
where $\text{Cov}(p^m) = \frac{1}{N - 1} \sum_m (p^m - \bar{p})(p^m - \bar{p})^T$. The final VicReg loss over the batch  is defined as:
\begin{equation}
    \begin{aligned}
    \mathcal{L}^{\text{\tiny VicReg}}_{p_i,z_j} & = \frac{\lambda}{n}\lVert p_i - z_j \rVert^2_2  + \mu \left[ v(P) + v(Z)\right] +  \nu \left[ c(P) + c(Z)\right].
    \end{aligned}
    \label{eq: loss_vicreg}
\end{equation}

\begin{table}[t]
\centering
\small
\setlength\tabcolsep{1.8pt}
\renewcommand{\arraystretch}{1.2}
\caption{\textbf{Semi-supervised evaluation on ImageNet.} We performed end-to-end finetuning using the settings in \ref{tab:impl_vicmae_finetune}, but disable RandAug and MixUp for this experiment.}
\label{tab:semi_sup_imagenet} 
% \vspace{-.5em}
\begin{tabular}{lcccccc}
\toprule
\textbf{Percentage of data} & \textbf{5\%} & \textbf{10\%} & \textbf{25\%} & \textbf{50\%} & \textbf{75\%} & \textbf{100\%} \\
\midrule
MAE \cite{he2022masked} + SimSiam \cite{chen2021exploring} & 7.15 & 23.41 & 39.73 & 54.94 & 62.88 & 67.44 \\
MAE \cite{he2022masked} + VicReg \cite{bardes2021vicreg} & 47.48 & 56.63 & 66.62 & 73.00 & 75.29 & 77.41 \\
\model (ours) & \textbf{50.25} & \textbf{58.22} & \textbf{67.65} & \textbf{73.97} & \textbf{75.80} & \textbf{77.89} \\
\bottomrule
\end{tabular}

\end{table}

We perform experiments using these two combinations of MAE and contrastive losses as baseline comparisons for our method but found them to be underperforming with only contrastive or only masked methods. In other words, it is not trivial to adapt constrastive learning methods to be used in combination with masked autoencoders. See Table \ref{tab:ours_vs_contrastive_loss} for more details.
For the contrastive learning part we experiment with two alternatives.
\begin{itemize}
    \item \textit{MAE + \{SimSiam or VicReg\}}. The predictor consists of the backbone network $f_{\theta}$ and a projector followed by a predictor as in Bardes~\etal~\cite{bardes2021vicreg}. The target encoder consists of the backbone $f_{\theta}$ and the projector, which are shared between the two encoders. 
    \item \model. The predictor and the target networks share the same architecture consisting of the backbone network $f_{\theta}$ and a projector following Bardes~\etal~\cite{bardes2021vicreg}.
\end{itemize}
When using the MAE + \{SimSiam or VicReg\} combinations, we use the $[\text{CLS}]$ token from the ViT architecture which is typically used to capture a global feature from the transformer network and is used to fine-tune the network for downstream tasks such as classification.

Combining MAE with negative-free representation learning is non trivial, and we set to test these by comparing our model with  MAE models with alternative negative-free learning objectives Siamsiam~\cite{chen2021exploring} and VicReg~\cite{bardes2021vicreg}. We present our results using linear evaluation on Table~\ref{tab:ours_vs_contrastive_loss}. We use the $[\text{CLS}]$ token as the global video representation for contrastive pre-training for 400 epochs. We can notice that competing methods underperform compared to our model which uses pooling of the local features by an absolute margin of $>3\%$ over the \mbox{\em MAE + VicReg} model.

\paragraph{Semi-supervised evaluation on ImageNet.}
\label{supp:semi_sup_imagenet}
We also test \model against negative-free representation learning methods on the problem of Semi-Supervised evaluation on the ImageNet dataset. The setting consists on training on a subset of the training data and testing on the whole validation data. We chose subsets of size 5\%, 10\%, 25\%, 50\%, 75\% and 100\% of the whole training set of ImageNet. We compare our model against MAE \cite{he2022masked} + SimSiam \cite{chen2021exploring}, and MAE \cite{he2022masked} + VicReg \cite{bardes2021vicreg}. Results are shown on Table \ref{tab:semi_sup_imagenet}, and show the supperiority of \model over simple combinations of contrastive learning and masked image modeling.

\section{Extra Ablations}
\noindent
\textbf{Temporal representation learning.} 
To investigate the temporal representation learned by our model, we conducted an ablation study on finetuned video models using the K400 dataset, examining performance through standard evaluation, frame shuffling, and frame repetition strategies. We evaluated and averaged 16 random permutations and exhaustive single-frame repetitions. Our findings reveal a decrease in accuracy from the original 87.8\% to 78.8\% with shuffled frames and 60\% with repeated frames, underscoring our model's ability to implicitly learn temporal representations despite not being explicitly designed for temporal modeling. These results demonstrate the model's effectiveness in finetuning for temporal representation learning, highlighting its capacity to capture diverse temporal features.
}
\end{document}